%% file: main.tex
\pdfoutput=1
\documentclass[11pt]{article}
\usepackage[preprint]{acl}
\usepackage{times}
\usepackage{latexsym}
\usepackage[T1]{fontenc}
\usepackage[utf8]{inputenc}
\usepackage{microtype}
\usepackage{inconsolata}
\usepackage{graphicx}
\usepackage{xspace}
\usepackage{color}
\usepackage{amsfonts}
\usepackage{amsmath}
\usepackage{booktabs}
\def\code#1{\texttt{#1}}
\newcommand{\icon}{\raisebox{-2pt}{\includegraphics[width=1.2em]{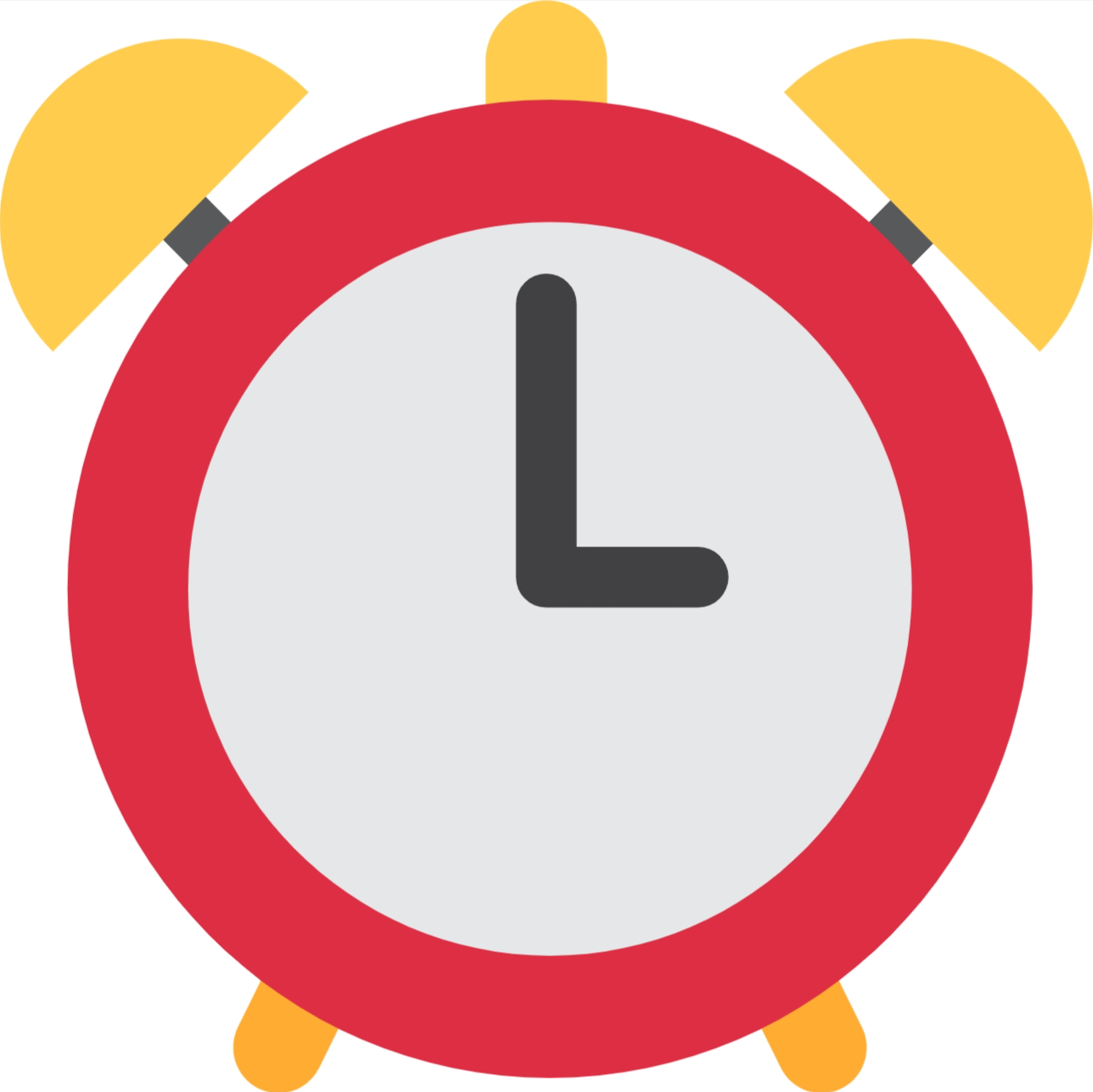}}\xspace}

\newcommand*{\affaddr}[1]{#1}

\title{\icon{} \textsc{ALaRM}: Align Language Models via Hierarchical Rewards Modeling}
\author{
{\normalsize Yuhang Lai$^\spadesuit$ ~ Siyuan Wang$^\clubsuit$ ~ Shujun Liu$^\clubsuit$\vspace{0.3mm} }\\ 
{\normalsize\bf Xuanjing Huang$^\diamondsuit$ ~ Zhongyu Wei$^{\clubsuit\triangle}$\footnotemark[2]\vspace{1mm}}\\
\affaddr{\normalsize$^\spadesuit$Institute of Science and Technology for Brain-inspired Intelligence, Fudan University\ \ } \vspace{0.2mm}\\
\affaddr{\normalsize$^\clubsuit$School of Data Science, Fudan University} \vspace{0.2mm}\\
\affaddr{\normalsize$^\diamondsuit$School of Computer Science, Fudan University} \vspace{0.2mm}\\
\affaddr{\normalsize$^\triangle$Research Institute of Intelligent and Complex Systems, Fudan University\ \ }\\
\normalsize\texttt{yhlai23@m.fudan.edu.cn, \{wangsy18,xjhuang,zywei\}@fudan.edu.cn}\\
}
\begin{document}
\maketitle
\footnotetext[2]{Corresponding author}
\input{abs}
\input{intro}
\input{framework}
\input{long-form_qa}
\input{mt}
\input{ablation}
\input{related-work}
\input{conclusion}
\input{limitations}
\input{ethics}
\bibliography{custom}
\clearpage
\newpage
\appendix
\input{appendix}
\end{document}

%% file: abs.tex
\begin{abstract} \label{sec:abs}
We introduce \textsc{ALaRM}, the first framework modeling hierarchical rewards in reinforcement learning from human feedback (RLHF), which is designed to enhance the alignment of large language models (LLMs) with human preferences. The framework addresses the limitations of current alignment approaches, which often struggle with the inconsistency and sparsity of human supervision signals, by integrating holistic rewards with aspect-specific rewards. This integration enables more precise and consistent guidance of language models towards desired outcomes, particularly in complex and open text generation tasks. By employing a methodology that filters and combines multiple rewards based on their consistency, the framework provides a reliable mechanism for improving model alignment. We validate our approach through applications in long-form question answering and machine translation tasks, employing \code{gpt-3.5-turbo} for pairwise comparisons, and demonstrate improvements over existing baselines. Our work underscores the effectiveness of hierarchical rewards modeling in refining LLM training processes for better human preference alignment.
We release our code at \url{https://ALaRM-fdu.github.io}.
\end{abstract}

%% file: intro.tex
\section{Introduction} \label{sec:intro}
\begin{figure}[tphb]
    \centering
    \includegraphics[width=0.91\columnwidth]{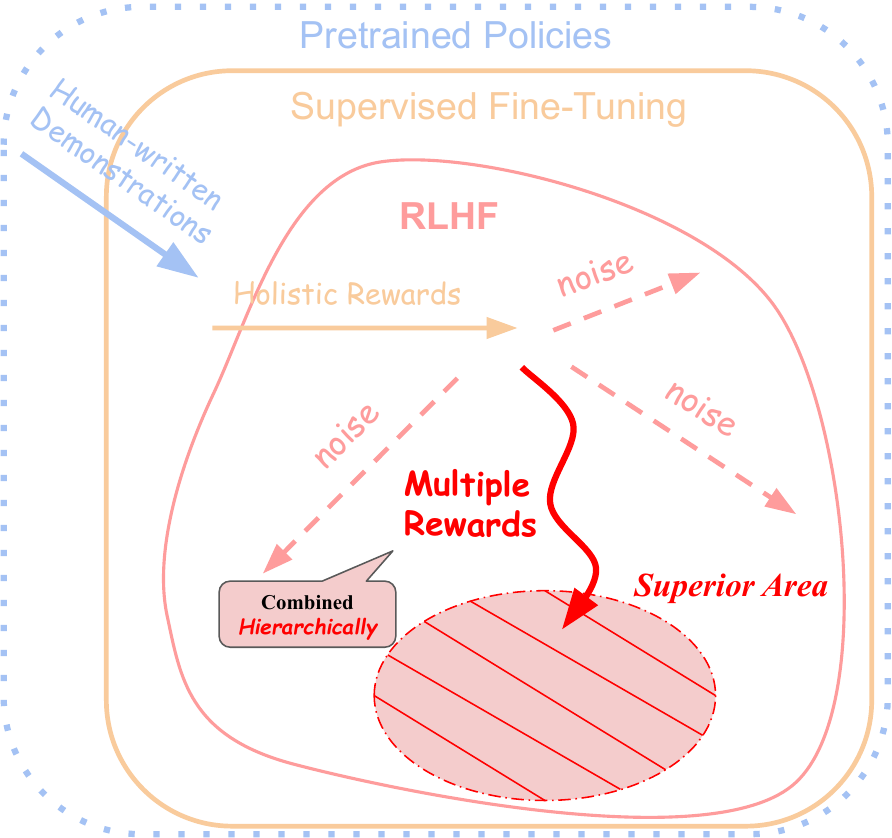}
    \caption{Illustration of our key ideas. The pretrained policies are first supervised fine-tuned on human-written demonstrations and then trained through RLHF given a holistic reward learned from human comparisons. The shadowed "superior area" better aligns with human preference, which is hard to reach for solely a noisy holistic reward. We propose to utilize multiple rewards hierarchically for more accurate and consistent supervision signals and thus guide the policies into the superior area.}
    \label{fig:training_signals}
\end{figure}
Current LLM-assisted AI systems have shown remarkable performance in a wide range of tasks \citep{gpt3, codex, llama2, llmagent}, and benefit from different forms of human supervision signals \citep{finetuned_lm, learningtosummarize}. While supervised learning relies on human-written demonstrations to unlock the emergent abilities gained from pretraining on huge text and code corpora, RLHF utilizes generation comparisons labeled by humans to further fine-tune the LLMs for better alignment with human expectations, which has been demonstrated to be able to reduce undesired model generations like harmful contents \citep{training_a_helpful} or hallucinations \citep{instructgpt}.

However, human oversight capabilities are finite.
As recent LLMs are capable of doing more complex work and even surpass human performance in some areas, it becomes more difficult to write good enough demonstrations even for human experts \cite{ds1000}.
Comparisons between several model generations can be intuitively easier to get from a crowd-sourcing platform though, previous research \citep{longeval, finegrained} has revealed that human annotations can be inconsistent and unreliable in evaluations between two or more model outputs for complex tasks like long text generations, producing unstable rewards in RLHF.
While stable rewards are the key to successful reinforcement learning, the sparsity nature of current holistic rewards further stresses this challenge \citep{drlc}.
Moreover, various scenarios hidden behind the tasks can differentiate preference standards \cite{psoups}, thus degrading the annotation consistency and value alignment on downstream applications, e.g., concise vs. comprehensive summary in text summarization, and literary vs. technical style in machine translation.
Then we ask the question, how to get reliable and scalable supervision signals within limited human oversight capabilities?

To take an initial step towards addressing this issue, we introduce a new framework \textsc{ALaRM} hierarchically modeling both holistic and aspect-specific rewards, which is motivated by 1) fine-grained RLHF \citep{finegrained} that categorizes different error types for more accurate and easier annotation, 2) 
task decomposition in hierarchical reinforcement learning \citep{HRL-survey} that helps to overcome sparse rewards.
At the core of our framework is to seek stronger supervision signals: As shown in \autoref{fig:training_signals}, solely using the holistic reward can make it difficult to reach the shadowed "superior area" which represents better alignment with human preference. Thus we employ multiple rewards combined in a hierarchical way to stabilize the optimization direction for more accurate and consistent guidance into the superior area.
Firstly, we list several aspect-specific rewards corresponding to the task and perform the selection by their inconsistency with the holistic reward in pairwise comparisons. Later in the RLHF training process, the chosen rewards are combined with the holistic reward as a whole once the sampled generation receives a high holistic reward, which is above a certain threshold value.
The aspect-specific rewards can either come from reward models trained on comparison datasets annotated along a specific dimension (e.g., honesty) or simply be toolkit-calculated metrics (e.g., token count), with an arbitrary density at either the token level or the sequence level. In addition, we proactively transform the aspect-specific rewards to guarantee that their cumulative values are positive, thereby motivating the policy to surpass the threshold for higher returns, which enriches the effectiveness of the hierarchical structure.

We apply our framework to two text generation tasks: long-form question answering (QA) and machine translation (MT). Long-form QA represents difficult annotations for complex tasks, and MT represents more flexible and various preference standards behind the tasks.
For each of them, we employ \code{gpt-3.5-turbo} as the evaluator for pairwise comparisons. We empirically demonstrate that our framework outperforms the compared baselines.
The ablation studies and corresponding analyses further show that our framework effectively provides stronger supervision signals toward human preference in both scenarios.

Collectively, we highlight our contributions as follows: 1) To our knowledge, we are the first to propose a framework that hierarchically models both holistic and aspect-specific rewards in RLHF, 2) we investigate how to perform reward selection to mitigate rewards conflicting, 3) we demonstrate the effectiveness of \textsc{ALaRM} as pursuing more accurate and consistent supervision signals on two text generation tasks through comprehensive ablation studies and analyses, shedding light on its potential for scalable oversight.

%% file: framework.tex
\begin{figure*}[tphb]
    \centering
    \includegraphics[width=0.97\textwidth]{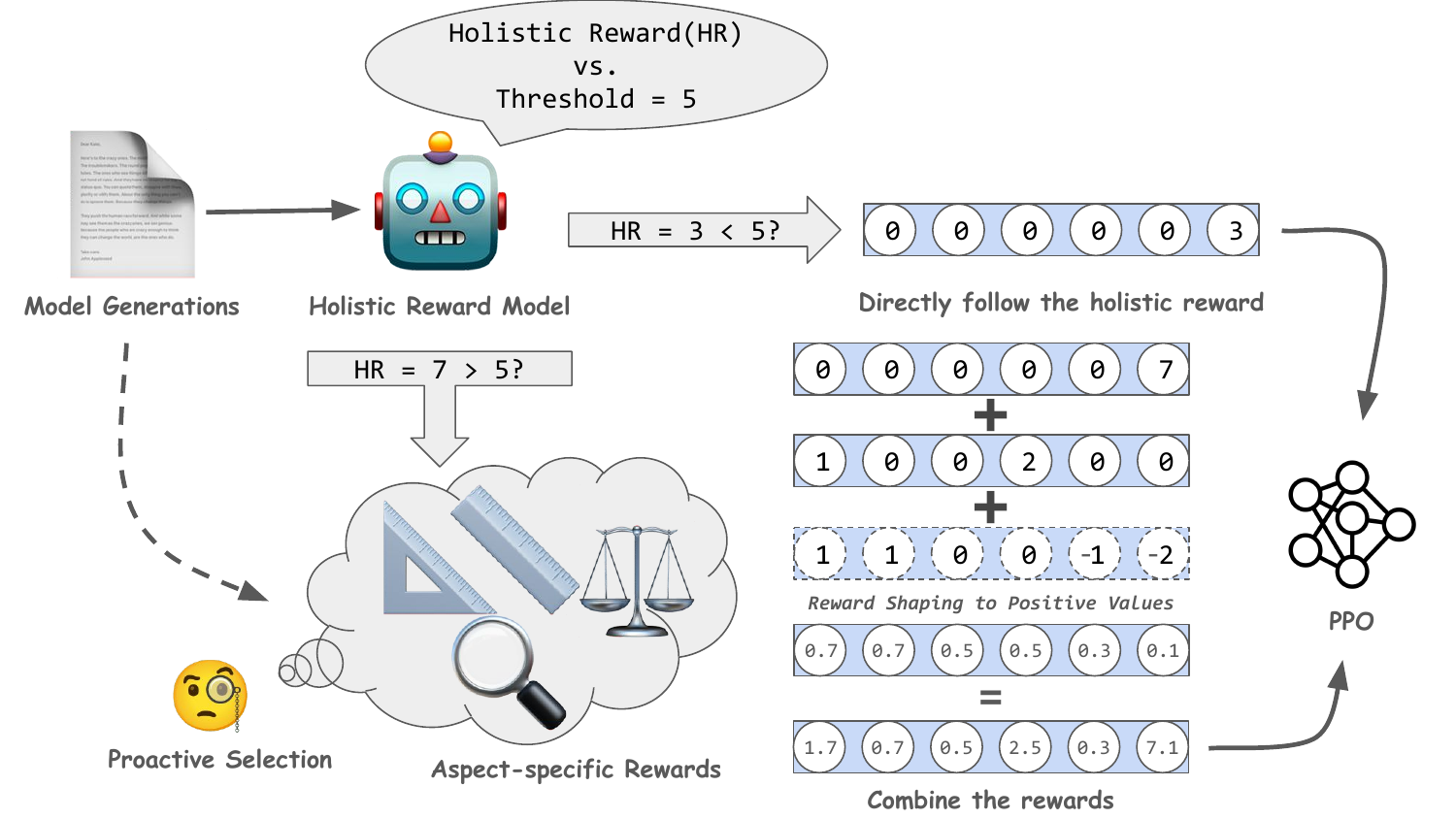}
    \caption{Illustration of our framework. The reward modeling is decomposed into two parts: 1) Directly assign the holistic reward to improve general quality, 2) combine the holistic reward and proactively selected aspect-specific rewards as a whole reward, which is supposed to be more accurate and consistent.}
    \label{fig:hierarchical}
\end{figure*}
\section{Framework} \label{sec:framework}
Our framework originates from RL-based text generation \citep{rl_summarization, rl_qa}, meanwhile modeling both holistic and aspect-specific rewards hierarchically. We first introduce the widely used RLHF. Then, we discuss why we should do reward selection proactively, and how we do that. After that, we present our method of hierarchical rewards modeling in detail.
\subsection{Reinforcement Learning from Human Feedback}
In the context of RL-based text generation, we define a language model parameterized by $\theta$ as a policy $\pi_{\theta}$. Token generation is considered a decision-making process, and the policy completes an episode when the language model generates an EOS token or meets the length limit.

In this way, RLHF aims to optimize the policy $\pi_{\theta}$ to maximize the reward from a preference model $R$ with Proximal Policy Optimization (PPO) \cite{ppo}, a broadly used reinforcement learning algorithm for human preference alignment. The preference model $R$ is learned on human-annotated comparison datasets to predict a single scalar that correctly ranks two or more model generations in the same comparison batch. The optimization objective is also formulated as follows:
\begin{equation}
\mathop{\arg\max}\limits_{\theta} \mathbb{E}_{\tau \sim \pi_{\theta}}[R(\tau)]
\end{equation}
where $\tau$ denotes a trajectory produced by the policy $\pi_{\theta}$, and $R(\tau)$ is the reward associated with the trajectory $\tau$, as evaluated by the preference model $R$. Here we see how the reward model $R$ deeply affects RLHF and thus it's crucial to ensure the accuracy and consistency of its prediction.
\subsection{Reward Selection}
Evaluation along a specific dimension of model generations instead of the general quality is demonstrated to be less noisy and more accurate for reward modeling \cite{finegrained}. Therefore, to get more accurate and consistent supervision signals, we first intuitively list several aspect-specific rewards corresponding to a certain task. However, human preferences are intricate. Different decomposed aspects are interconnected and can even conflict with each other. A common way to balance them is the weighted sum method, which assigns a carefully chosen weight for each aspect-specific reward, based on observations of either performance during training or accuracy in pairwise comparisons \cite{compositional}. Nevertheless, this method still suffers from the over-optimization problem \cite{confronting}, where the model loses individual information from every single aspect-specific reward and cannot attribute changes in the composed reward to any of them. 

Our key idea, which differs from merely combining all aspect-specific rewards, is to stabilize the supervision signals. We need a "copilot" for the holistic reward. Thus we aim to resolve this challenge by discarding the conflicting rewards, and we select the rewards that are mostly consistent with the holistic reward. Therefore, we proactively conduct pairwise comparisons between two sets of model generations. These generations come from the same supervised fine-tuned model but are produced using greedy decoding and pure sampling, respectively. Then, we calculate the prediction inconsistency by assessing whether the holistic and aspect-specific rewards have divergent predictions on which answer is better.

\subsection{Hierarchical Rewards Modeling}
Hierarchical reinforcement learning has advanced significantly in a wide range of decision-making tasks \citep{HRLforActionControl, HRLforDialog, HRLforVideoCaption}, decomposing complex and challenging optimization objectives into simpler sub-tasks. Nevertheless, in contrast, existing RLHF works typically employ a plain rewarding strategy that linearly assigns a single holistic reward \citep{alignLMMfactual} or a fixed combination of aspect-specific rewards \citep{compositional}, which not only poses sparse rewards in the long-horizon optimization but also overlooks the close relationships between the holistic reward and the aspect-specific rewards.

With these motivations, we propose a novel approach that leverages both holistic and aspect-specific rewards. In this way, we consider the optimization objective that aligns the language models with human preference, targeting the superior area depicted in \autoref{fig:training_signals}, as a challenging decision-making task. Thus we propose a decomposition of this task into two less complex sub-tasks which ought to be addressed sequentially: 1) Directly follow the holistic reward until the model generation receives a high holistic reward, which indicates the generation is generally good and meets human preference at a relatively high level, 2) optimize the combination of the holistic and aspect-specific rewards, which as a whole provides more accurate and consistent supervision signals towards the superior area. Unlike the plain weighted-sum method, which applies combined rewards throughout the entire training, our approach is more nuanced. We primarily follow the supervision of the holistic reward and gently turn the steering wheel only when it's insufficient to solely rely on the holistic reward to reach the superior area.

As illustrated in \autoref{fig:hierarchical}, we first follow the regular RLHF process to sample some generations from the policy. Then we employ a preference model that predicts a single scalar as the holistic reward for each of them. Here we utilize UltraRM-13B \cite{ultrarm} as a zero-shot reward model to predict holistic rewards for all experiments reported in this paper.
For each sampled generation, if it receives a holistic reward lower than a certain threshold value, we directly assign it as the final reward. Otherwise, we calculate the proactively selected aspect-specific rewards and combine all the rewards together.
To design a more effective hierarchical architecture for the above reward modeling, we ensure that the generations receiving a holistic reward above the threshold obtain higher cumulative rewards than those below this threshold. We achieve this through reward shaping, where we transform aspect-specific rewards into positive values using the sigmoid function.

%% file: long-form_qa.tex
\section{Long-Form Question Answering} \label{sec:long-form qa}
\begin{figure}[tphb]
    \centering
    \includegraphics[width=0.85\columnwidth]{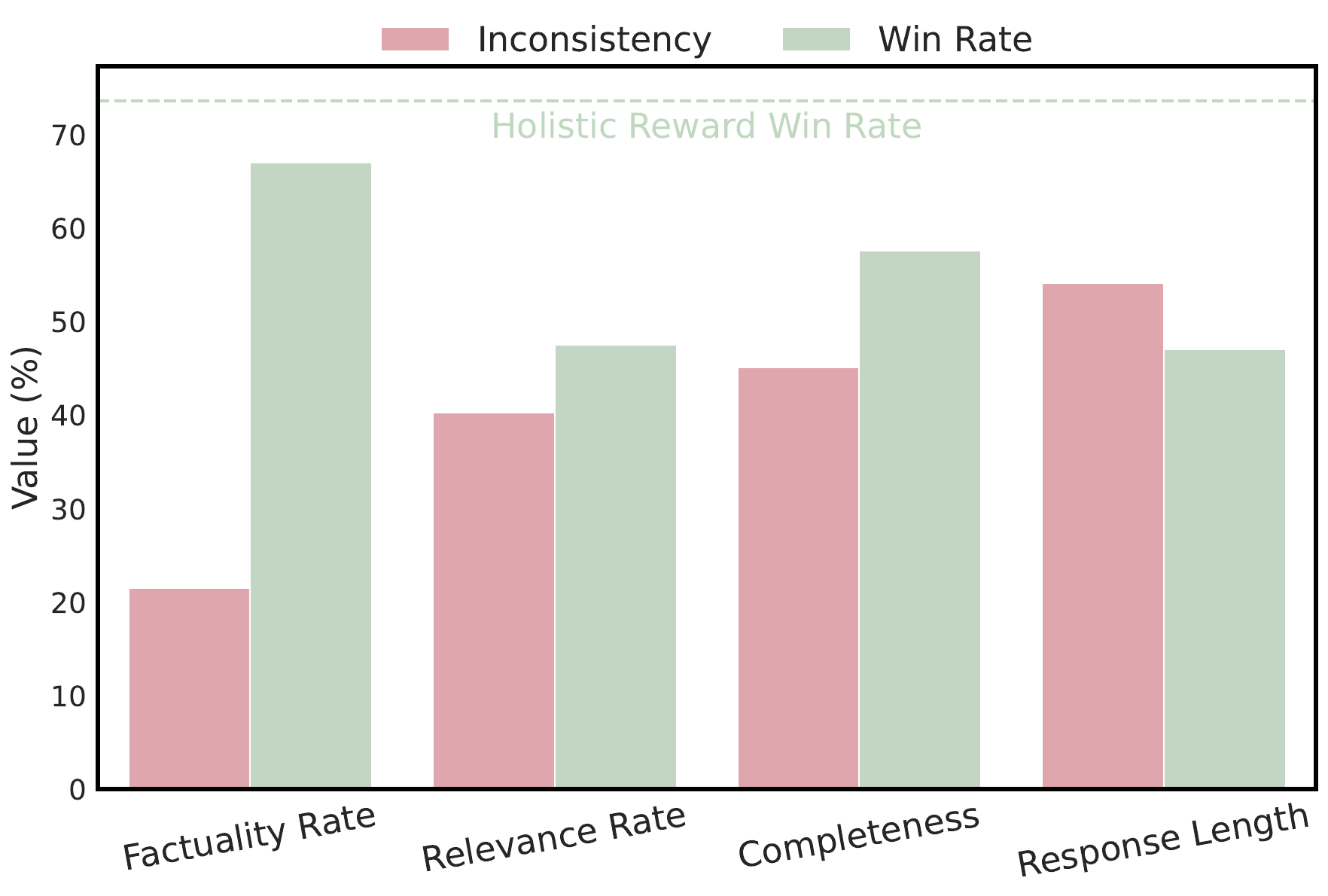}
    \caption{Inconsistency with the holistic reward for listed aspect-specific rewards and win rates of the greedy decoding against the pure sampling in long-form QA.}
    \label{fig:LFQA-reward-selection}
\end{figure}
Long-form QA is a complicated task that aims to produce elaborate responses covering background information, explanations, or discussions corresponding to the questions. This process is difficult and even humans are not able to write high-quality demonstrations or make accurate and consistent comparisons reliably. Following we show how we address this issue using \textsc{ALaRM}.
\subsection{Task Settings}
We utilize most of the settings and release from \citet{finegrained}: QA-Feedback dataset, supervised fine-tuned T5-large as the initial policy, and three fine-grained reward models.
\paragraph{Dataset.}QA-Feedback dataset is extracted from ASQA \cite{asqa} and then transformed into a task format of reading comprehension, which gives an ambiguous factoid question and a set of related knowledge sources from the Wikipedia corpus, and requires the language model to generate a long-form response. QA-Feedback has a training set of 3,853 examples, a development set of 500, and a test set of 948 in total. And we keep the division of this dataset in the experiment.
\input{LFQA-win-rate}
\input{LFQA-reward-mean}
\paragraph{Initial Policy.}The initial policy model is T5-large \citep{T5} initialized with supervised fine-tuning on 1K training examples.
\paragraph{Reward Models.}We reuse the three fine-grained reward models designed in \citet{finegrained}, which are called the relevance reward model $R_{\phi_1}$, the factuality reward model $R_{\phi_2}$, the completeness reward model $R_{\phi_3}$, representing three categories of different error types. They all use Longformer-base \citep{longformer} as the backbone model and do predictions at different levels of density. $R_{\phi_1}$ is trained to predict whether the generation contains irrelevance, repetition, or incoherence errors at the sub-sentence level, with a binary classification accuracy of 69.6 and an F1-score of 68.5 on the development set. $R_{\phi_2}$ learns to detect incorrect or unverified facts for each sentence according to given knowledge sources, having an accuracy of 77.8 and an F1-score of 67.5. $R_{\phi_3}$ is developed to measure the holistic information completeness for the full sequence, with an accuracy of 70.9 in pairwise comparisons.
\subsection{Reward Selection} \label{subsec:LFQA-reward-selection}
\paragraph{Listing Corresponding Rewards.}Following the task settings, we have three aspect-specific reward models. We also consider the response token count as one corresponding reward in this task.
\paragraph{Calculating Inconsistency}To proactively filter out appropriate rewards that mostly aid the holistic reward, together as more accurate and consistent training signals, we conduct pairwise comparisons to check the inconsistency with the holistic reward of those four candidate rewards.

We first employ the initial policy to get two sets of model generations on the training set, using greedy decoding and pure sampling respectively, and two generations from different sets to the same question form a comparison pair. For each aspect-specific reward, we calculate the inconsistency as the percentage of the pairs that the holistic reward prefers one, while the aspect-specific reward prefers another. We disregard all the ties that have the same holistic reward. For the relevance reward and the factuality reward, they cannot make an apple-to-apple comparison due to their sentence or sub-sentence level reward density. Thus we compute the overall correct rate in all the slices of each generation for direct comparison, named factuality rate and relevance rate.

As shown in \autoref{fig:LFQA-reward-selection}, the factuality rate shows significantly lower inconsistency than the other three rewards, indicating it is more suitable to be the "copilot" of the holistic reward. Also, we compared the win rates of the greedy decoding generations to the pure sampling ones. The win rate is calculated as described in \autoref{subsec:experimental-setup}. Note that the factuality rate has more similar win rates to the holistic reward, which further supports its consistency. Therefore, we select the factuality reward for hierarchical rewards modeling.
\subsection{Experimental Setup} \label{subsec:experimental-setup}
\paragraph{Rewards Modeling.}We z-normalize the holistic reward in a set of generations $D_p$ which is produced through pure sampling on the training set. The factuality reward is shaped to positive values to ensure the hierarchical structure by the sigmoid function. The threshold value is set at 0.6, which is around the top 30\% in $D_p$. We simply add the rewards up to combine them and only adjust the weight of the holistic reward.
\paragraph{Reinforcement Learning Training.}We use the pure sampling strategy in the reinforcement learning process and use greedy decoding for the development set and the test set evaluation. Beginning with the supervised fine-tuned initial policy, we train our model for 2 epochs on the training set. As we set the exploration frequency to 4, the training runs for about 30K episodes. We utilize LoRA \citep{lora} for the training.
\paragraph{Evaluation.}Besides the mean value of the holistic reward and the factuality rate, we evaluate the win rates between different models in pairwise comparisons. Following \citet{psoups}, the win rate is formulated as:
\begin{equation}
     \textit{Win~Rate} = \frac{\textit{Win}}{\textit{Win} + \textit{Lose}}   
\end{equation}
where all ties are disregarded in the calculation. We also evaluate the win rates in general quality by employing \code{gpt-3.5-turbo} as the evaluator for pairwise comparison. To mitigate the positional bias \citep{positional-bias} in LLM-as-a-judge, we perform prompting twice with the two generations swapped. We only consider the comparison valid when both promptings prefer the same generation, indicating the evaluator is faithful enough to overcome the positional bias. We use the prompt (\autoref{fig:prompt}) from AlpacaEval \citep{alpacaeval}.
\paragraph{Compared Methods.}We compare our framework to three methods. \textbf{\textsc{ALaRM}} represents our proposed hierarchical rewards modeling approach. \textbf{Holistic Reward} represents the baseline using the holistic reward as the only supervision signal. \textbf{Factuality Reward} represents solely using the factuality reward without reward shaping to train the policy. \textbf{Weighted Sum} is the plain weighted sum method that directly adds the holistic reward and the factuality reward together for training.

\subsection{Main Results}
In this paper, we conduct all experiments using three different seeds, and the results are averaged across these three independent runs. \autoref{tab:LFQA-reward-mean} shows the evaluation results on the test set of the mean values of each reward. We can see that \textsc{ALaRM} leads to significantly higher holistic reward than other methods, meanwhile reaching the highest factuality rate. As expected, except for \textsc{ALaRM}, Holistic Reward gets the highest holistic reward value and Factuality Reward gets the highest factuality rate. Weighted Sum balances these two rewards instead. \autoref{tab:LFQA-win-rate} represents the win rates between the four methods. And we can see \textsc{ALaRM} holds the best under all three different metrics, which further indicates that \textsc{ALaRM} provides a stronger supervision signal than other methods.

%% file: LFQA-win-rate.tex
\begin{table*}[!t]
\centering
\resizebox{0.9\linewidth}{!}{
\begin{tabular}{l|cccc|c}
\toprule
Methods & Holistic Reward & Factuality Reward & Weighted Sum & \textsc{ALaRM} & Avg.\\ \midrule
 \multicolumn{6}{c}{\textit{\% Win Rates by the Holistic Reward}} \\ \midrule
Holistic Reward & - & $55.86\pm4.22$ & $50.36\pm0.95$ & $49.06\pm0.48$ & $51.76\pm1.66$\\
Factuality Reward  & $44.13\pm4.22$ & - & $45.04\pm3.73$ &$42.73\pm3.91$ & $43.97\pm3.92$ \\
Weighted Sum & $49.64\pm0.95$ & $54.95\pm3.73$ & - & $48.55\pm1.14$ & $51.05\pm0.67$\\ \midrule
\textsc{ALaRM} & $\textbf{50.94}\pm0.48$ & $\textbf{57.27}\pm3.91$ & $\textbf{51.45}\pm1.14$ & - & $\textbf{53.22}\pm1.76$\\ \midrule
 \multicolumn{6}{c}{\textit{\% Win Rates by the Factuality Rate}} \\ \midrule
Holistic Reward  & - & $48.31\pm3.49$ & $50.89\pm1.50$ &$44.42\pm2.58$ & $47.87\pm1.48$ \\
Factuality Reward  & $51.68\pm3.49$ & - & $52.76\pm4.01$ &$48.70\pm3.17$ & $51.05\pm3.49$ \\
Weighted Sum & $49.11\pm1.50$ & $47.24\pm4.01$ & - & $44.65\pm4.40$ & $47.00\pm 3.00$\\ \midrule
\textsc{ALaRM} & $\textbf{55.58}\pm2.58$ & $\textbf{51.30}\pm3.17$ & $\textbf{55.35}\pm4.40$ & - & $\textbf{54.08}\pm2.15$\\ \midrule
 \multicolumn{6}{c}{\textit{\% Win Rates Evaluated by \code{gpt-3.5-turbo}}} \\ \midrule
Holistic Reward  & - & $44.11\pm3.12$ & $44.16\pm6.28$ &$40.94\pm1.87$ & $43.07\pm1.81$ \\
Factuality Reward  & $55.88\pm3.12$ & - & $53.55\pm6.09$ &$50.77\pm4.01$ & $53.40\pm4.21$ \\
Weighted Sum & $55.84\pm6.28$ & $46.45\pm6.09$ & - & $46.03\pm5.94$ & $49.44\pm5.93$\\ \midrule
\textsc{ALaRM} & $\textbf{59.06}\pm1.87$ & $\textbf{49.23}\pm4.01$ & $\textbf{53.96}\pm5.94$ & - & $\textbf{54.08}\pm2.18$\\
\bottomrule
\end{tabular}
}
\caption{Evaluation of win rates determined by the holistic reward, the factuality rate, and the evaluator \code{gpt-3.5-turbo} respectively in long-form QA.}
\label{tab:LFQA-win-rate}
\end{table*}

%% file: LFQA-reward-mean.tex
\begin{table}[hptb]
\centering
\resizebox{0.9\linewidth}{!}{
\begin{tabular}{l|ccc}
\toprule
Methods & HR($\uparrow$) & FR($\uparrow$) & Length \\ \midrule
Holistic Reward  & $0.608$ & $0.736$ & $87.8$\\
Factuality Reward & $0.538$ & $0.752$ & $116.8$\\
Weighted Sum & $0.598$ & $0.738$ & $98.5$\\ \midrule
\textsc{ALaRM}  & $\textbf{0.617}$ & $\textbf{0.752}$ & $97.6$\\
\bottomrule
\end{tabular}
}
\caption{Evaluation results on the test set in long-form QA. Here HR represents the mean of the holistic reward. FR represents the mean of the factuality rate. Length represents the mean token count of each generation.}
\label{tab:LFQA-reward-mean}
\end{table}

%% file: mt.tex
\section{Machine Translation} \label{sec:mt}
Machine translation can be considered a text generation task that involves converting a piece of text from the source to the target language while preserving the original meaning, context, and cultural nuances. This task requires not only a deep understanding of the grammatical structure and vocabulary of both the source and target languages but also an appreciation of their idiomatic expressions and cultural references. However, this feature can differentiate people's preferences and form noise \citep{readability, gender_bias}, due to the wide range of application scenarios and user groups, thus posing barriers to accurate and consistent alignment annotation. Below, we show how we use \textsc{ALaRM} on this task.
\input{MT-win-rate}
\input{MT-reward-mean}
\subsection{Task Settings}
\paragraph{Dataset.}We utilize Europarl \citep{europarl}, a Spanish-English dataset that contains transcripts of European Parliamentary proceedings. We select 100K samples from it and arrange them into a training set of 63K, a development set of 2K, and a test set of 30K in total.
\begin{figure}[!b]
    \centering
    \includegraphics[width=0.85\columnwidth]{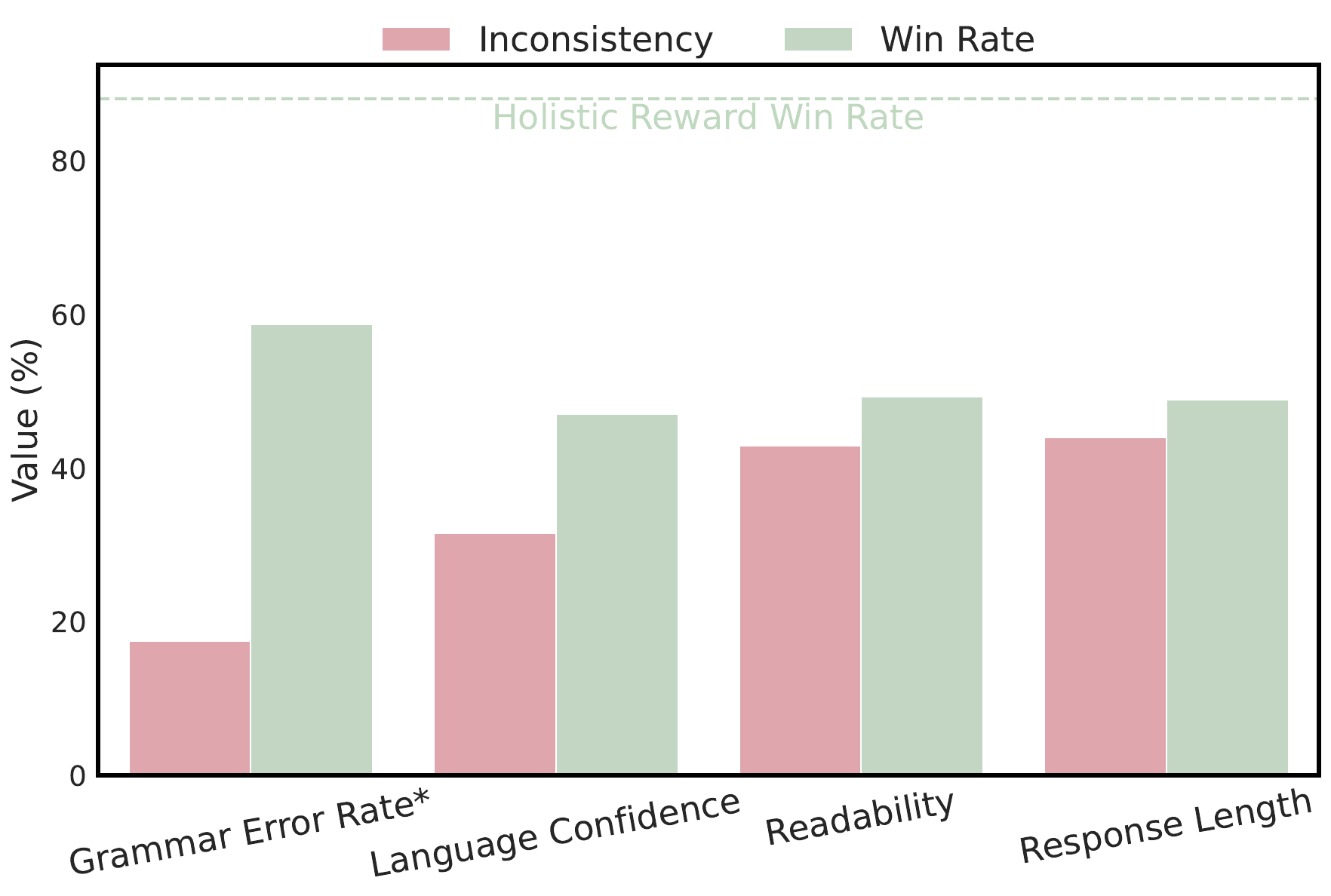}
    \caption{Selection results of inconsistency and win rates in MT. ${}^{*}$: The lower grammar error rate wins.}
    \label{fig:MT-reward-selection}
\end{figure}
\paragraph{Initial Policy.}We initialize the policy with mT5-base \citep{mt5}, and then supervised fine-tuned it on the training set for 5 epochs. The initial policy has a BLEU score of 31.57 on the test set.
\subsection{Reward Selection}
\paragraph{Listing Corresponding Rewards.}We first intuitively list three corresponding rewards, named grammar reward, language confidence, and readability. All of them are calculated through python-wrapped toolkits. The grammar reward utilizes LanguageTool, which can detect grammar errors at the word level. We define the grammar reward as assigning negative values to the grammatically incorrect tokens. The language confidence is based on Lingua, which computes the language likelihood in a set of languages by n-gram models as a single scalar. The readability is to measure the general difficulty of reading the text, which can be calculated by Textstat at the full sequence level.
\paragraph{Calculating Consistency.}The same as in \autoref{subsec:LFQA-reward-selection}, we conduct pairwise comparisons on two sets of generations produced by greedy decoding and pure sampling respectively. We define the grammar error rate of a generation as the token count divided by the number of grammar errors.

As shown in \autoref{fig:MT-reward-selection}, the grammar reward stands out with lower inconsistency and better win rates than other rewards. Thus we choose the grammar reward for the following experiments.
\subsection{Experimental Setup}
We follow most setups in \autoref{subsec:experimental-setup}. Evaluation and compared methods are the same.
\paragraph{Rewards Modeling.}We apply z-normalization to the holistic reward and set the threshold value at 0.5, which corresponds to the top 30\%. Considering the token level density of the grammar reward, we apply reward shaping for every token, including those with a 0 reward, using the sigmoid function to maintain the hierarchy.
\paragraph{Reinforcement Learning Training.}With the pure sampling strategy in training, we train our models for around 13K episodes.
\subsection{Main Results}
\autoref{tab:MT-reward-mean} represents the mean values of separate rewards on the test set, where \textsc{ALaRM} has the highest holistic reward and a comparable grammar error rate to the other methods. As shown in \autoref{tab:MT-win-rate}, \textsc{ALaRM} continues to outperform the others in the win rates, evaluated by the holistic reward, the grammar error rate, and \code{gpt-3.5-turbo} respectively. These results together strongly support the effectiveness of our framework on this task.

%% file: MT-win-rate.tex
\begin{table*}[tphb]
\centering
\resizebox{0.9\linewidth}{!}{
\begin{tabular}{l|cccc|c}
\toprule
Methods & Holistic Reward & Grammar Reward & Weighted Sum & \textsc{ALaRM} & Avg.\\ \midrule
 \multicolumn{6}{c}{\textit{\% Win Rates by the Holistic Reward}} \\ \midrule
Holistic Reward & - & $51.52\pm0.77$ & $50.18\pm0.89$ & $48.30\pm0.66$ & $50.00\pm0.76$\\
Grammar Reward  & $48.47\pm0.77$ & - & $48.69\pm0.15$ &$47.07\pm0.16$ & $48.08\pm0.32$ \\
Weighted Sum & $49.82\pm0.89$ & $51.31\pm0.15$ & - & $48.26\pm0.15$ & $49.80\pm0.25$\\ \midrule
\textsc{ALaRM} & $\textbf{51.70}\pm0.66$ & $\textbf{52.93}\pm0.16$ & $\textbf{51.74}\pm0.15$ & - & $\textbf{52.12}\pm0.26$\\ \midrule

 \multicolumn{6}{c}{\textit{\% Win Rates by the Grammar Error Rate$^{1}$}} \\ \midrule
Holistic Reward  & - & $51.90\pm2.83$ & $53.47\pm3.58$ & $42.20\pm4.29$ & $49.19\pm0.40$ \\
Grammar Reward  & $48.10\pm2.83$ & - & $50.63\pm5.11$ & $42.47\pm5.29$ & $47.07\pm3.82$ \\
Weighted Sum & $46.54\pm3.58$ & $49.37\pm5.11$ & - & $40.63\pm4.58$ & $45.51\pm 3.88$\\ \midrule
\textsc{ALaRM} & $\textbf{57.80}\pm4.29$ & $\textbf{57.53}\pm5.29$ & $\textbf{59.37}\pm4.58$ & - & $\textbf{58.24}\pm4.46$\\ \midrule

 \multicolumn{6}{c}{\textit{\% Win Rates$^{2}$ Evaluated by \code{gpt-3.5-turbo}}} \\ \midrule
Holistic Reward  & - & $51.03\pm0.42$ & $51.27\pm1.79$ &$48.97\pm1.70$ & $50.43\pm1.00$ \\
Grammar Reward  & $48.97\pm0.42$ & - & $48.98\pm2.77$ &$47.85\pm2.09$ & $48.60\pm0.90$ \\
Weighted Sum & $48.73\pm1.79$ & $51.02\pm2.77$ & - & $47.98\pm3.05$ & $49.24\pm2.30$\\ \midrule
\textsc{ALaRM} & $\textbf{51.03}\pm1.70$ & $\textbf{52.15}\pm2.09$ & $\textbf{52.02}\pm3.05$ & - & $\textbf{51.73}\pm2.11$\\
\bottomrule
\end{tabular}
}

\caption{Evaluation of win rates in MT. ${}^{1}$: The lower grammar error rate wins in the calculation. ${}^{2}$: We choose a smaller set of 3K examples randomly selected from the test set to reduce the annotation cost.}
\label{tab:MT-win-rate}
\end{table*}

%% file: MT-reward-mean.tex
\begin{table}[tphb]
\centering
\resizebox{0.95\linewidth}{!}{
\begin{tabular}{l|ccc}
\toprule
Methods & HR($\uparrow$) & \% GER($\downarrow$) & Length \\ \midrule
Holistic Reward  & $0.919$ & $1.203$ & $36.7$\\
Grammar Reward & $0.908$ & $\textbf{1.190}$ & $36.5$\\
Weighted Sum & $0.918$ & $1.197$ & $36.5$\\ \midrule
\textsc{ALaRM}  & $\textbf{0.928}$ & $1.203$ & $37.0$\\
\bottomrule
\end{tabular}
}
\caption{The means of rewards on the test set in MT. GER represents the mean grammar error rate.}
\label{tab:MT-reward-mean}
\end{table}

%% file: ablation.tex
\section{Ablation Study} \label{sec:ablation}
\input{Ablation-reward-selection}
\paragraph{Without Selection.}As shown in \autoref{tab:ablation-reward-selection}, to find out how reward selection affects the performance of \textsc{ALaRM}, we conduct extensive experiments that separately apply each reward listed in the initial pool on both tasks. The proactively selected rewards present leading performance evaluated by both the holistic reward and \code{gpt-3.5-turbo}, showing the effectiveness of reward selection. We also observe conflicting scores for some rewards from the two evaluators. We attribute this to the biases and flaws in the holistic reward, such as constantly overlooking or overvaluing certain aspects \cite{llm_judge}, a concern that differs from the inconsistency issues we focus on and falls outside this paper's scope. See full tables in \autoref{appendix:experimental_details}.
\paragraph{Without Combination.}To examine whether \textsc{ALaRM} helps with more accurate and consistent supervision signals by utilizing both holistic and aspect-specific rewards, we compare the methods using separate rewards individually in the experiments. As shown in Table \ref{tab:LFQA-win-rate} and \ref{tab:MT-win-rate}, \textsc{ALaRM} consistently leads to better results along both dimensions.
\paragraph{Without Hierarchical Structure.}We contrast our framework with the conventional weighted sum method to highlight the significance of the hierarchical structure. The results from the weighted sum approach reflect a compromise between holistic and aspect-specific rewards, limiting its ability to excel in both. Conversely, our framework, \textsc{ALaRM} leverages hierarchical rewards modeling to provide more potent supervision signals, enhancing its performance in both dimensions.

%% file: Ablation-reward-selection.tex
\begin{table}[t]
\centering
\resizebox{\linewidth}{!}{
\begin{tabular}{lcc}
\toprule
\textbf{Rewards} & \multicolumn{2}{c}{\textbf{Avg. Win Rates Against Others by}} \\
\cmidrule{2-3} & Holistic Reward & \code{gpt-3.5-turbo}\\
\midrule
 \multicolumn{3}{c}{\textit{Task: Long-Form Question Answering}} \\
\textbf{Factuality} & $\textbf{51.93}\pm1.66$ & $\underline{54.74}\pm0.70$ \\
Relevance & $46.89\pm0.72$ & $\textbf{55.44}\pm1.12$ \\
Completeness & $49.51\pm0.82$ & $44.97\pm1.52$ \\
Length & $\underline{51.67}\pm0.22$ & $44.84\pm1.32$ \\
\midrule
 \multicolumn{3}{c}{\textit{Task: Machine Translation$^{*}$}} \\
\textbf{Grammar} & $\textbf{51.58}\pm0.64$ & $\textbf{50.61}\pm1.62$ \\
Confidence & $49.20\pm0.46$ & $48.66\pm1.26$ \\
Readability & $49.53\pm0.42$ & $\underline{50.59}\pm1.06$ \\
Length & $\underline{49.69}\pm0.26$ & $50.14\pm1.75$ \\
\bottomrule
\end{tabular}
}
\caption{Averaged win rates against others on both tasks. Each reward is used in \textsc{ALaRM} separately. ${}^{*}$: Evaluation by \code{gpt-3.5-turbo} is on a smaller set of 3K randomly selected examples as in \autoref{tab:MT-win-rate}.}
\label{tab:ablation-reward-selection}
\end{table}

%% file: related-work.tex
\section{Related Work} \label{sec:related-work}
\paragraph{Hierarchical Reinforcement Learning.}Designing a hierarchical rewarding structure that decomposes a long-horizon reinforcement learning task into simpler sub-tasks has shown promising performance in traditional reinforcement learning problems \citep{hrl_p1,hrl_p2,hrl_p3,hrl_p4}. Motivated by this, \textsc{ALaRM} first utilizes hierarchical reinforcement learning to align language models.
\paragraph{Human Preference Alignment.}AI alignment with human preference has been one of the key research topics in the NLP community as LLMs show notable performance yet are prone to generating unexpected content \citep{human_preference_align1,human_preference_align2,human_preference_align3,human_preference_align4}. RLHF is a popular algorithm for AI alignment and many related methods are proposed \citep{rlhf_related1,rlhf_related2,rlhf_related3}.
\paragraph{Scalable Oversight.}Superhuman models should be capable of handling complex and creative tasks beyond human expertise \citep{weaktostrong}, which raises the increasingly important issue of scalable oversight: how to provide reliable supervision signals within limited human capabilities \citep{amodei2016concrete,bowman2022measuring}? Current methods include improving evaluation quality through human-AI collaboration \citep{irving2018ai,christiano2018supervising} and simplifying tasks into sub-tasks for more reliable assessments \citep{wu2021recursively,zhong2023nonprogrammers,lightman2023lets,cosupervised}. Our framework adopts the latter one, identifying simpler aspects for evaluation.

%% file: conclusion.tex
\section{Conclusion} \label{sec:conclusion}
We propose \textsc{ALaRM}, the first framework hierarchically modeling both holistic and aspect-specific rewards in RLHF. We explore proactive reward selection strategies to enhance compatibility with the holistic reward. The effectiveness of our framework in seeking more accurate and consistent supervision signals and its potential to inspire scalable oversight in AI alignment, is demonstrated through comprehensive experiments, detailed ablation studies, and analyses across two text generation tasks.

%% file: limitations.tex
\section*{Limitations}
Our framework requires rewards that are specifically designed for each task, which poses challenges in scaling up the application scenarios. We need to improve the automatic selection of rewards. In our evaluation, we utilize OpenAI's API, which incurs additional costs and may experience rate limitations and unstable response times for regular users.

%% file: ethics.tex
\section*{Ethics Statement}
Our study does not involve direct human or animal subjects and presents no discernible ethical concerns. The datasets used, such as QA-Feedback, Europarl, and the toolkits, including Textstat, Lingua, and LanguageTool, are publicly available. We have taken steps to ensure transparency and reproducibility in our research. We confirm that our research and methodologies are free from harmful practices and potential misuse. We are committed to upholding the highest standards of integrity and ethical responsibility in our work.

%% file: appendix.tex
\section{Experimental Details} \label{appendix:experimental_details}
\begin{figure}[tphb]
    \centering
    \includegraphics[width=0.7\linewidth]{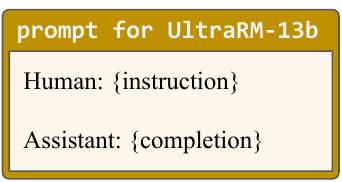}
    \caption{The inference prompt for UltraRM.}
    \label{fig:ultra_prompt}
\end{figure}

We adopt the TRL implementation for the PPO algorithm and conduct all training using LoRA and BF16. The prompt for inferencing UltraRM is shown in \autoref{fig:ultra_prompt}. We list the resources and materials used in this paper in \autoref{tab:resource_materials}.
\input{appendix-resource}
\subsection{Training Details on Long-form QA}
We limit the max input length to 1024 and the max output length to 200 following the original setting. We set the weight of the holistic reward to $5$ through trials and keep other weights to $1$. The factuality reward model predicts $0.5$ and $-0.5$ for the correct and wrong sentences respectively. The batch size is set to 8 and the mini-batch size to 4, with gradient accumulation steps set to 2. We set the learning rate to $2e-4$. Other PPO hyper-parameters should follow the default values of TRL implementation. The training takes around 2 hours on 8×48G NVIDIA A6000 GPU.
\subsection{Training Details on MT}
We set the max length to 128 for both input and output. We set the weight of the holistic reward to $3$ through trials and keep other weights to $1$. The grammar reward is designed to predict $-1$ for those incorrect tokens and 0 for the correct ones before the reward shaping. The batch size is set to 32 and the mini batch size is set to 16, with gradient accumulation steps set to 2. We set the learning rate to $5e-4$. Other PPO hyper-parameters also follow the default values. The training takes around 1 hour on 8×48G NVIDIA A6000 GPU.

\subsection{Additional Ablation Study Results on Reward Selection} \label{appendix:additional_ablation_results}
\input{appendix-ablation-LFQA-reward-selection}
\input{appendix-ablation-MT-reward-selection}
We put the full results of win rates in the ablation study for reward selection in \autoref{tab:appendix-additional-ablation-LFQA-reward-selection} and \autoref{tab:appendix-additional-ablation-MT-reward-selection}. For those extensive experiments, We use the same settings as the main ones. The relevance model predicts $0.3$ and $-0.3$ for the correct and wrong sub-sentences respectively. We apply z-normalization to the readability and completeness rewards. We divide the token count of a generation by the average token count on the training set and define the value as its length reward. The language confidence remains the same for rewarding since it has a suitable value range between $0$ and $1$.

\section{Evaluation Details}
\begin{figure*}[t]
    \centering
    \includegraphics[width=0.9\textwidth]{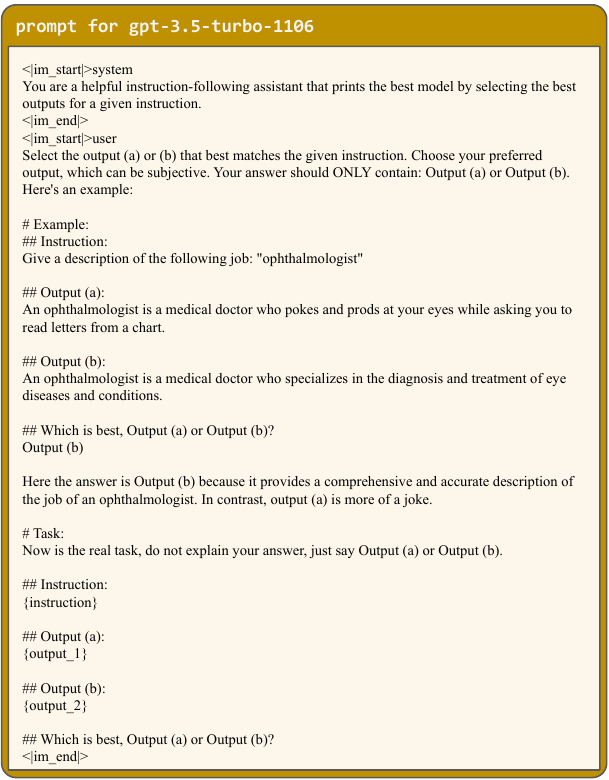}
    \caption{The evaluation prompt for \code{gpt-3.5-turbo-1106}.}
    \label{fig:prompt}
\end{figure*}

We use the \code{gpt-3.5-turbo-1106} version of OpenAI API for pairwise evaluation. \autoref{fig:prompt} shows our prompt to call \code{gpt-3.5-turbo} in pairwise comparisons for both tasks, which is originally from AlpacaEval \citep{alpacaeval}.

%% file: appendix-resource.tex
\begin{table*}[hptb]
    \centering
\resizebox{0.9\linewidth}{!}{
\begin{tabular}{cc}
\toprule
Resources and Materials & Access Link \\ \midrule
UltraRM-13b & \url{https://huggingface.co/openbmb/UltraRM-13b} \\
Long-form QA & \url{https://github.com/allenai/FineGrainedRLHF}\\
Europarl & \url{https://huggingface.co/datasets/Helsinki-NLP/europarl}  \\
LanguageTool   & \url{https://github.com/languagetool-org/languagetool}  \\
Lingua   & \url{https://github.com/pemistahl/lingua-py}  \\
textstat   & \url{https://github.com/textstat/textstat}  \\
TRL & \url{https://github.com/huggingface/trl} \\
\bottomrule
\end{tabular}
}
\caption{The resources and materials we utilize in this paper.}
\label{tab:resource_materials}
\end{table*}

%% file: appendix-ablation-LFQA-reward-selection.tex
\begin{table*}[hptb]
\centering
\resizebox{0.95\linewidth}{!}{
\begin{tabular}{l|cccc|c}
\toprule
Rewards & Factuality & Relevance & Completeness & Length & Avg.\\ \midrule
 \multicolumn{6}{c}{\textit{\% Win Rates by the Holistic Reward}} \\ \midrule
Factuality & - & $53.57\pm2.06$ & $51.74\pm1.64$ & $50.49\pm1.58$ & $\textbf{51.93}\pm1.66$\\ \midrule
Relevance  & $46.43\pm2.06$ & - & $47.88\pm0.49$ &$46.36\pm0.30$ & $46.89\pm0.72$ \\
Completeness & $48.26\pm1.64$ & $52.12\pm0.49$ & - & $48.16\pm1.33$ & $49.51\pm0.82$\\
Length & $49.51\pm1.58$ & $53.64\pm0.30$ & $51.83\pm1.33$ & - & $\underline{51.67}\pm0.22$\\
\midrule
 \multicolumn{6}{c}{\textit{\% Win Rates Evaluated by \code{gpt-3.5-turbo}}} \\ \midrule
Factuality & - & $48.95\pm1.01$ & $56.57\pm1.89$ & $58.70\pm0.75$ & $\underline{54.74}\pm0.70$\\ \midrule
Relevance  & $51.05\pm1.01$ & - & $58.33\pm0.64$ &$56.96\pm1.79$ & $\textbf{55.44}\pm1.12$ \\
Completeness & $43.43\pm1.89$ & $41.67\pm0.64$ & - & $49.81\pm2.44$ & $44.97\pm1.52$\\
Length & $41.30\pm0.75$ & $43.04\pm1.79$ & $50.19\pm2.44$ & - & $44.84\pm1.32$\\
\bottomrule
\end{tabular}
}
\caption{Additional ablation study results of win rates in long-form QA.}
\label{tab:appendix-additional-ablation-LFQA-reward-selection}
\end{table*}

%% file: appendix-ablation-MT-reward-selection.tex
\begin{table*}[hptb]
\centering
\resizebox{0.95\linewidth}{!}{
\begin{tabular}{l|cccc|c}
\toprule
Rewards & Grammar & Confidence & Readability & Length & Avg.\\ \midrule
 \multicolumn{6}{c}{\textit{\% Win Rates by the Holistic Reward}} \\ \midrule
Grammar & - & $51.79\pm0.85$ & $51.60\pm0.78$ & $51.36\pm0.30$ & $\textbf{51.58}\pm0.64$\\ \midrule
Confidence  & $48.21\pm0.85$ & - & $49.79\pm0.15$ &$49.60\pm0.44$ & $49.20\pm0.46$ \\
Readability & $48.40\pm0.78$ & $50.21\pm0.15$ & - & $49.98\pm0.57$ & $49.53\pm0.42$\\
Length & $48.64\pm0.30$ & $50.40\pm0.44$ & $50.02\pm0.57$ & - & $\underline{49.69}\pm0.26$\\
\midrule
 \multicolumn{6}{c}{\textit{\% Win Rates$^{*}$ Evaluated by \code{gpt-3.5-turbo}}} \\ \midrule
Grammar & - & $51.31\pm2.11$ & $50.19\pm0.44$ & $50.34\pm2.50$ & $\textbf{50.61}\pm1.62$\\ \midrule
Confidence  & $48.69\pm2.11$ & - & $48.41\pm1.71$ & $48.87\pm1.19$ & $48.66\pm1.26$ \\
Readability & $49.81\pm0.44$ & $51.59\pm1.71$ & - & $50.38\pm1.95$ & $\underline{50.59}\pm1.06$\\
Length & $49.66\pm2.50$ & $51.13\pm1.19$ & $49.62\pm1.95$ & - & $50.14\pm1.75$\\
\bottomrule
\end{tabular}
}
\caption{Additional ablation study results of win rates in MT. ${}^{*}$: Evaluation is on a smaller set of 3K randomly selected examples as in \autoref{tab:MT-win-rate}.}
\label{tab:appendix-additional-ablation-MT-reward-selection}
\end{table*}